\documentclass[letterpaper, 10 pt, conference]{IEEEtran}  %

\IEEEoverridecommandlockouts                              %

\usepackage{graphicx}        %
\usepackage{multicol}        %
\usepackage[bottom]{footmisc}%
\usepackage[capbesideposition=outside,capbesidesep=quad]{floatrow}
\usepackage{hyperref}
\usepackage[export]{adjustbox}
\usepackage{cite}
\hypersetup{bookmarksopen,bookmarksnumbered,
pdfpagemode=UseOutlines,
colorlinks=true,
linkcolor=blue,
anchorcolor=blue,
citecolor=blue,
filecolor=blue,
menucolor=blue,
urlcolor=blue
}
\usepackage{marvosym}

\usepackage[T1]{fontenc}    %
\usepackage{upgreek}
\usepackage{amsfonts,amssymb}
\usepackage{bm}
\usepackage{bbm}

\usepackage{amsthm} %
\usepackage{thmtools}
\usepackage{mathtools}
\usepackage{epstopdf}
\usepackage{xspace}
\usepackage{outlines}
\usepackage[]{caption} %
\usepackage[font=footnotesize]{subcaption}
\usepackage{units}
\usepackage{colortbl}
\usepackage{tabulary}
\usepackage[usenames,dvipsnames]{xcolor} %
\usepackage{diagbox}
\usepackage[ruled,vlined,linesnumbered,noend]{algorithm2e}  %
\usepackage{parskip}
\SetKwComment{Comment}{$\triangleright$\ }{}
\usepackage{ifthen,version}
\usepackage{soul}
\usepackage{fixltx2e}
\usepackage{csquotes}
\usepackage{microtype}      %
\usepackage{lipsum}
\usepackage{csquotes}
\usepackage{tikz}
\let\labelindent\relax
\usepackage{enumitem}
\usepackage{wrapfig}

\usepackage{multirow}
\usepackage{pbox}

\SetCommentSty{mycommfont}

\SetKwInput{KwInput}{Input}                %
\SetKwInput{KwOutput}{Output}              %

\newcommand{\xxnote}[3]{}
\ifx\hidenotes\undefined
  \usepackage{color}
  \renewcommand{\xxnote}[3]{\color{#2}{#1: #3}}
\fi

\newtheoremstyle{hypstyle}
{3pt} %
{3pt} %
{\itshape} %
{} %
{\bfseries} %
{.} %
{.5em} %
{} %

\theoremstyle{hypstyle}

\DeclareMathOperator*{\argmin}{arg\,min}

\newcommand{\real}[0]{\mathbb{R}}

\newcommand{\bbm}{\begin{bmatrix}}
\newcommand{\ebm}{\end{bmatrix}}

\newcommand{\library}{\bm{\mathcal{L}}}

\newcommand{\Traj}[0]{\xi}
\newcommand{\TrajRef}[0]{\Traj_{\text{ref}}}
\newcommand{\Data}[1]{\mathcal{D}_{#1}}

\newcommand{\tube}{\Omega}

\newcommand{\xv}{\mathbf{x}}
\newcommand{\uv}{\mathbf{u}}

\newcommand{\dv}{\mathbf{d}}
\newcommand{\gv}{\mathbf{g}}
\newcommand{\fv}{\mathbf{f}}

\newcommand{\FreeSpace}{\mathcal{X}_{free}}
\graphicspath{{figs/}}

\title{\LARGE \bf Adaptive Safety Margin Estimation for Safe Real-Time Replanning \\ under Time-Varying Disturbance
\vspace{-2mm}}

\author{Cherie Ho, Jay Patrikar, Rogerio Bonatti, Sebastian Scherer\\[0.5mm]
Carnegie Mellon University
\vspace{1.88mm}

}

\author{Cherie Ho$^{1}$, Jay Patrikar$^{1}$, Rogerio Bonatti$^{1}$, Sebastian Scherer$^{1}$%
\thanks{$^{1}$The Robotics Institute, Carnegie Mellon University \{\tt\small cherieh, jpatrika, rbonatti, basti\}@cs.cmu.edu}%
}

\begin{document}

\maketitle
\thispagestyle{empty}
\pagestyle{empty}

\begin{abstract}
Safe navigation in real time is challenging because engineers need to work with uncertain vehicle dynamics, variable external disturbances, and imperfect controllers.
A common safety strategy is to inflate obstacles by hand-defined margins. 
However, arbitrary static margins often fail in more dynamic scenarios, and using worst-case assumptions is overly conservative for most settings where disturbances over time.
In this work, we propose a middle ground: safety margins that \textit{adapt} \textit{on-the-fly}.
In an offline phase, we use Monte Carlo simulations to pre-compute a library of safety margins for multiple levels of disturbance uncertainties.
Then, at runtime, our system estimates the current disturbance level to query the associated safety margins that best trades off safety and performance.
We validate our approach with extensive simulated and real-world flight tests. 
We show that our \textit{adaptive} method significantly outperforms static margins, allowing the vehicle to operate up to 1.5 times faster than worst-case static margins while maintaining safety.\\
Video: \url{https://youtu.be/SHzKHSUjdUU}

\end{abstract}

\begin{figure}[h!]
     \centering
     \begin{subfigure}[b]{0.9\textwidth}
         \centering
         \includegraphics[width=\textwidth]{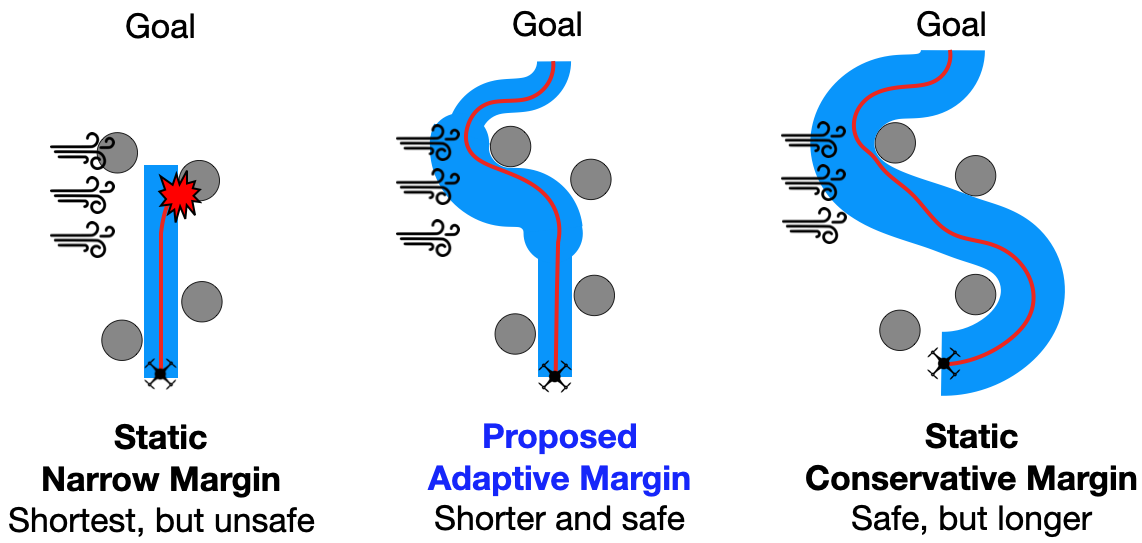}
         \caption{Concept}
     \end{subfigure}
     \\
     \begin{subfigure}[b]{0.9\textwidth}
         \centering
         \includegraphics[width=\textwidth]{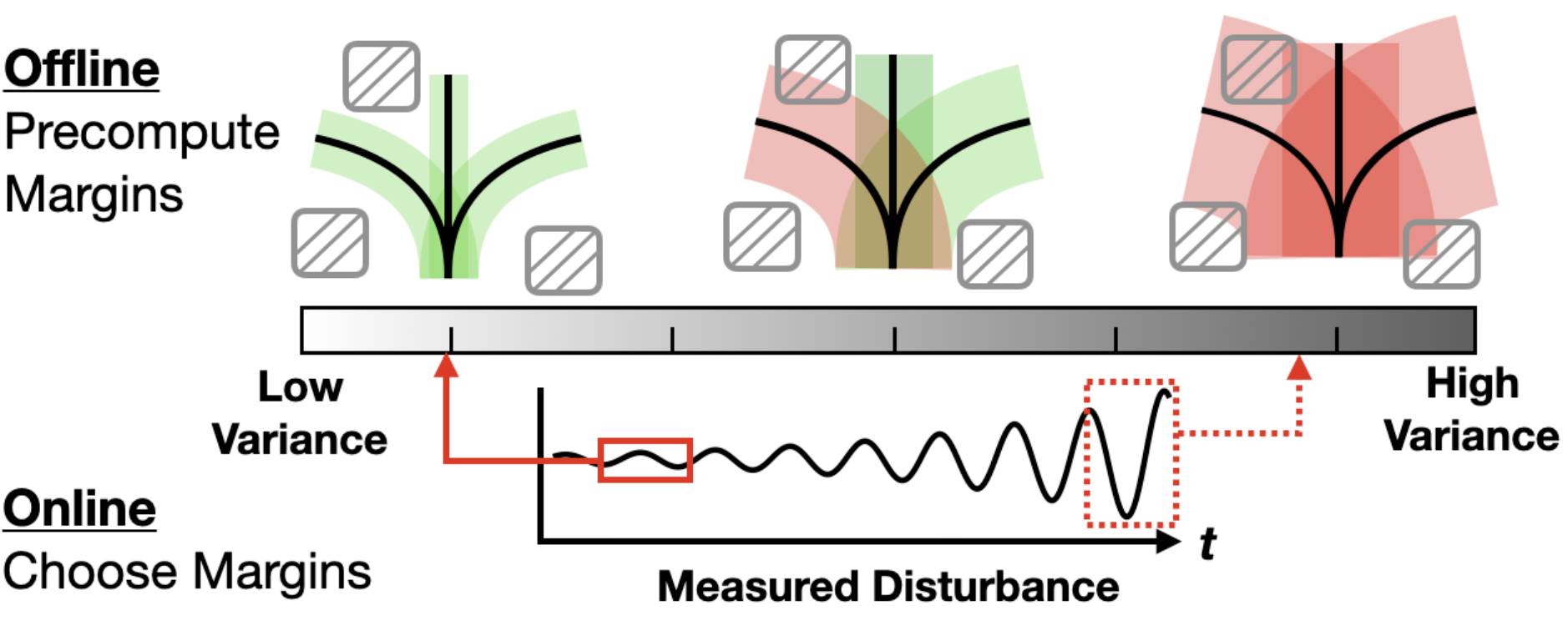}
         \caption{Approach}
     \end{subfigure}
     \\
     \begin{subfigure}[b]{0.9\textwidth}
         \centering
         \includegraphics[width=\textwidth]{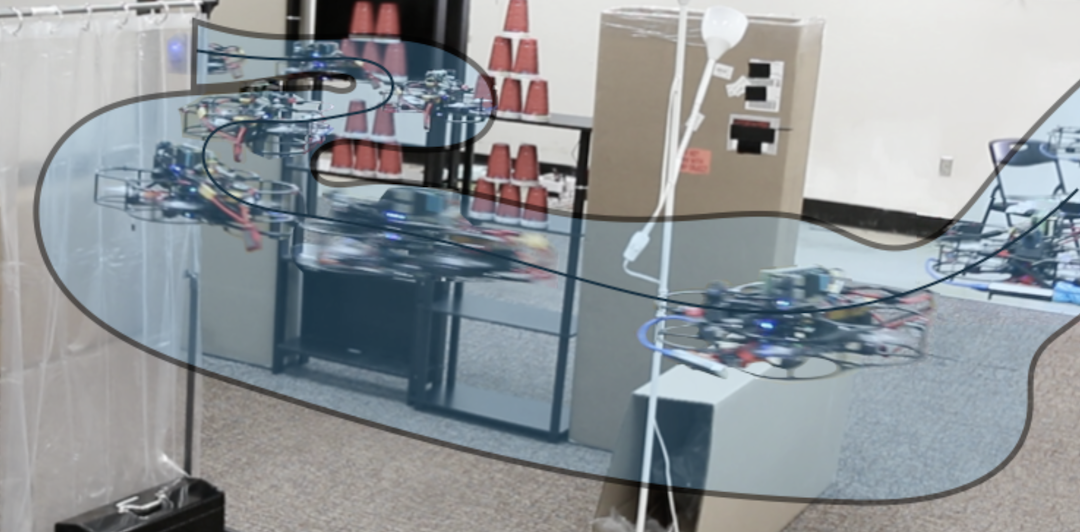}
         \caption{Execution}
     \end{subfigure}
     \caption{\small \textit{(a)} Given the current disturbance information, our proposed framework \textit{adapts} safety margins \textit{on-the-fly} to compute safe plans without overly sacrificing performance. \textit{(b)} Real-time adaptation is enabled by precomputing margins for a motion primitive library under multiple disturbance variances. \textit{(c)} We validate our method with flights in tight spaces and under time-varying disturbances.}
     \label{fig:key-figure}
\end{figure}
\section{Introduction}
\label{sec:intro}

In the past decades, robotics applications have been increasingly shifting from structured domains, where robots operate with full knowledge of all environmental variables, towards deployment in unstructured scenarios. 
We now find robots performing tasks as diverse as cargo delivery \cite{choudhury2019efficient}, security \cite{de2018ethics} and even cinematography \cite{bonatti2019autonomous}.
Much of this progress stems from advancements in perception and planning algorithms, as well as hardware and compute capacity.
However, ensuring autonomous vehicle safety still remains a central challenge, especially in dynamic navigation settings.

During navigation, robots must cope with disturbance uncertainties and remain safe among obstacles. Engineers often employ simplified vehicle dynamics and external disturbance models to handle the high-dimensionality of kinodynamic planning, leaving the burden of trajectory-following for the controller.
Consequently, the robot trajectory will display tracking errors, which can eventually lead to crashes despite a nominal collision-free path.
As seen in Fig.~\ref{fig:key-figure}, our work aims to address this challenge by computing probabilistically-safe trajectories despite model and disturbance uncertainties.

A traditional strategy to achieve safety is to inflate obstacle borders, or to craft safety tubes around trajectories.
Human experts hand-tune static safety margins for a particular mission based on previous experiences and costly trial-and-error \cite{gao2019flying,spitzer2018fast}. However, this approach works only under low variability of vehicle dynamics and environmental disturbances. A designer can also derive margins given a fixed bounded uncertainty, however this is not a realistic assumption in real-life situations. For example, we can use worst-case margins to account for large disturbances \cite{herbert2017fastrack, fridovich-keil_planning_2018, majumdar_funnel_2017, althoff_online_2015}, but worst-case guarantees lead to overly-conservative plans and lower navigation performance. Such static paradigm may either be unsafe or overly-conservative for missions where disturbance magnitude is \textit{time-varying} (e.g. transition between low-wind to high-wind regions).

Our work presents a middle ground between handcrafted and overly-conservative margins, capable of achieving good navigation performance without sacrificing safety. We develop \textit{adaptive} safety margins, which are selected using online disturbance measurements.
The key contribution is the introduction of \textit{adaptive} margins to motion primitive library planning approaches that enables better safety/performance tradeoff under time-varying disturbances.
Our approach encompasses uncertainties in the robot's dynamics model, controller and environmental disturbances, and can additionally operate under time-varying regimes. 
Our contributions are threefold:

\begin{enumerate}
    \item \textbf{Probabilistic Safety Tube Formulation: }First, we formalize the theory behind the design of probabilistically-safe tubes. (Section~\ref{sec:prob_def});
    \item \textbf{Real-time Adaptive Margins: } Next, we develop an offline-online approach to probabilistically guarantee vehicle safety. In the offline step, we employ our formal safety definition to pre-compute a library of safety margins at varying levels of disturbance uncertainties for a library of motion primitives. Online, we leverage the library of margins and use disturbance measurements to select the least conservative motion primitive that still guarantees safety for real-time planning. The proposed framework is invariant to choice of motion primitives, global planner, vehicle dynamics, controller and underlying disturbance distribution. (Section~\ref{sec:approach});
    \item \textbf{Experimental Validation: }Finally, we validate our real-time adaptive approach using extensive simulated and real-world flight tests among obstacles and under influence of time-varying disturbances. We show that our method has better performance than static margin baselines, while remaining safe (Section~\ref{sec:experiments}).
\end{enumerate}

\section{Related Work}
\label{sec:related_work}

\textbf{Planning under Uncertainty: } 
Obstacle inflation is often used to achieve safe planning under uncertainty \cite{liu2018search}.  In practice, many works hand-define static safety buffers \cite{spitzer2018fastteleop, gao2019flying,patrikar2020windplan} to generate safe trajectories, however an expert designer has to opaquely adjust margin sizes to balance between safety and performance. Majority of the path planning work also does not explicitly address external disturbance \cite{liu2018search,li2015sparse,lan2016bit,xie2015toward,allen2016real}.

\textbf{Robust Motion Planning: }We find works that precompute reachable sets given a high bounded disturbances to guarantee safety, such as, static funnels around a motion primitive library \cite{majumdar_funnel_2017}, a worst-case tracking error bound \cite{herbert2017fastrack,fridovich-keil_planning_2018}, or a robust control invariant sets for emergency maneuvers \cite{althoff_online_2015}. Another work derives a globally invariant tube that is valid for any dynamically feasible trajectory \cite{singh_robust_2019}.  However, above works do not adjust their disturbance bounds during a mission.

Kousik et. al \cite{kousik2019safe} learns a tracking error function offline to conservatively approximate reachable set at runtime, with hardware implementation and time-varying uncertainties as future work.
Most similar to our problem space, \cite{seo2019robust} approximates in real-time the reachable set given current disturbance, while our work precomputes margins for faster adaptation.  

\textbf{Robust, Adaptive Control: } Alternatively, model mismatch and obstacle avoidance can be cohesively handled in the control layer \cite{lopez2019dynamic}. Many of these approaches use a form of disturbance estimator/predictor \cite{ostafew_lnmpc_2016} to compensate for learnt model mismatch, however do not handle safety when model is uncertain. We find works \cite{ostafew_robustLNMPC_2016, fan2019bayesian} that explicitly account for model uncertainty to stay probabilistically-safe, gaining higher performance with experience, however plan for a short time horizon due to high computation cost.

We highlight that this paper does not propose a new planning method, but serves as a framework to estimate safety margins for motion primitive library methods given current disturbance to maintain probabilistic safety under time-varying disturbance. We take inspiration from disturbance estimation, adaptive control and robust motion planning to introduce adaptive probabilistically-safe margins for trajectory planning.

\section{Problem Definition}
\label{sec:prob_def}

We wish to control a robot under unknown varying disturbance to stay probabilistically safe while navigating as close as possible to a reference trajectory from a high-level planner. Our objective is to select, from a motion primitive library, the most similar primitive to reference trajectory such that when followed by the robot, it will remain safe under current disturbance given a risk threshold.

To do so, we estimate safety tubes that probabilistically enclose the nominal robot states at discretized disturbance uncertainties. We define a trajectory as probabilistically safe when its associated tube is not in collision.

\begin{figure*}[t]
  \center
  \includegraphics[width=1.0\textwidth]{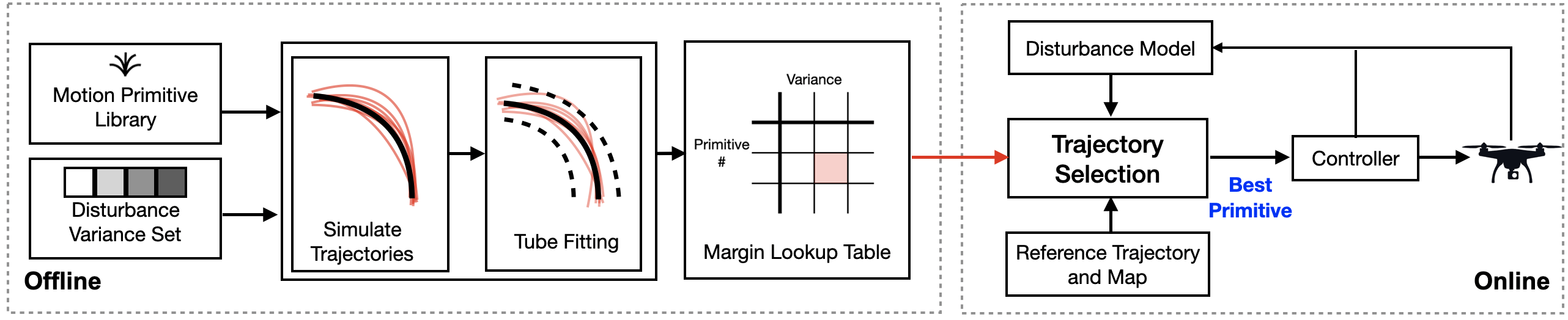}
  \caption{\small Overall Approach: \textit{Offline}, we generate a lookup table (LUT) of safety margins for each primitive in a motion primitive library at discretized levels of disturbance variances. \textit{Online}, given a map and a reference trajectory, 
  our system selects the best collision-free primitive within the library, with collision checks based on LUT margins using estimated disturbance variance. 
  \vspace{-4.0mm}}
  \label{fig:approach}
\end{figure*}
\subsection{Defining Trajectory Safety}
 Let $\xv \in \real^n$ be the robot state, and $\pi:(\xv, \Traj, t) \mapsto \uv \in \real^m$ be a controller that maps the current state $\xv$, trajectory $\Traj$, and time $t$ to a control command. We consider a robot described by a dynamics function $\fv_{true}$, which we define as the summation of a known prior model $\hat{\fv}$ and a disturbance  $\dv$. 
\begin{equation}
    \dot{\xv}(t) = \fv_{true}\big(\xv(t),\uv(t)\big) = \hat{\fv}\big(\xv(t), \uv(t)\big) + \dv(t)
\end{equation}
Let $\Traj:[0,t_f]\rightarrow \real^3 \times SO(2)$, be a trajectory followed by the vehicle, i.e., $\Traj(t) = \{x(t), y(t), z(t) , \psi(t)\}$. We define trajectory safety as a confidence bound $1-\epsilon$ that the vehicle remains in free space ($\FreeSpace\subseteq \real^3$) when following a trajectory $\Traj$, subject to disturbance $\dv$ and a controller $\pi$.
The time-varying disturbance captures the \textit{model mismatch} between the nominal model and the true vehicle dynamics. 

\textbf{Definition 1 (Probabilistic Trajectory Safety):}
\begin{align}
    &\forall t\in[0,t_f],~P(\xv(t) \in \mathbf{X}_{free}) \geq 1-\epsilon : \label{eq:prob_traj}\\
    &\dot{\xv}(t) = \fv_{true}\big(\xv(t),\uv(t)\big), ~~\uv(t) = \pi\big(\xv(t), \Traj(t)\big)\nonumber
\end{align}

Our trajectory safety definition plays an important practical and theoretical role. As described in Sec.~\ref{sec:intro}, operators commonly handle robot safety by inflating obstacles using handtuned margins. This process can be viewed as an instance of our safety tube definition, with tube width chosen by the expert before a mission. In contrast, our definition offers a more principled method for defining probabilistically safe trajectories, as we estimate the optimal tube's shape in real-time.

\begin{figure}[h]
    \centering
    \includegraphics[width=0.7\textwidth]{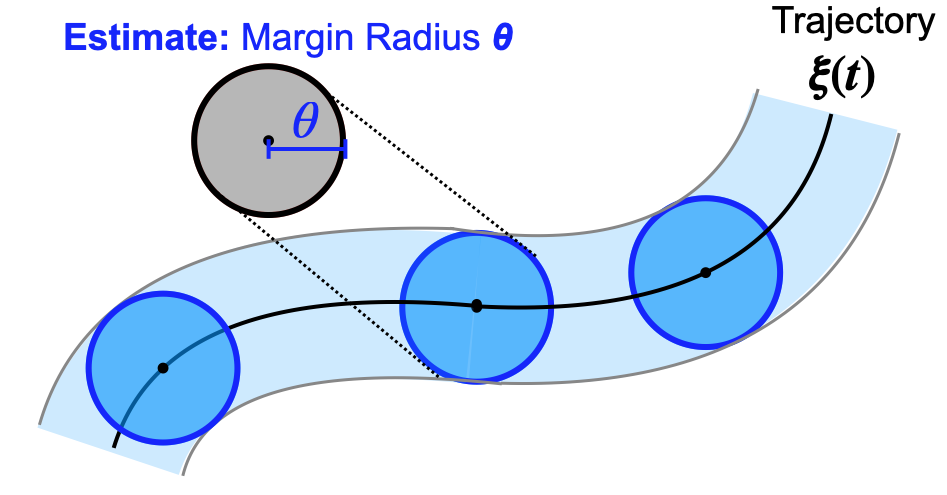}
    \caption{\small Tube $\tube_\Traj$ Definition. Given a trajectory $\Traj$ and current disturbance, we wish to estimate the tube's margin radius $\theta$.
    \vspace{-4.0mm}}
    \label{fig:tube_def}
\end{figure}

\subsection{Estimating Probabilistically-Safe Tube}
Our key idea is to compute a probabilistic safety margin for Eq.~\ref{eq:prob_traj} in terms of trajectory tracking error. First, we define the concept of a 3D safety tube $\tube_{\Traj}$, relative to a trajectory $\Traj$. Given an additive disturbance function $\dv: t \rightarrow \dv \in \real^n$ and a controller $\pi$, we define $\tube_{\Traj}$ as the minimum volume in workspace that encloses the vehicle states when tracking $\Traj$, with a confidence bound $1-\epsilon$.
We can describe the shape of the tube $\tube_{\Traj} \in \real^3$ with a \textit{margin radius} $\sigma$ that defines a 2D cross-section circle $S_t$ that is centered and perpendicular to the trajectory $\Traj$.
\begin{equation}
    ~~\tube_{\Traj} = \{\xv \in S_t~|~t\in[0,t_f]\} 
    \label{eq:tube_def}
    \vspace{-10pt}
\end{equation}

\subsection{Replanning Objective with Safety Tubes}
Our objective is to navigate as close as possible to a reference trajectory $\TrajRef$, given a similarity metric function $J_{sim}:(\Traj, \Traj_{ref})\rightarrow \real$. The reference trajectory can be user-defined or from a high-level planner, for example, a segment between start and goal locations, or a segment from a sampling-based planner. We do not assume that $\TrajRef$ is inherently collision-free (\textit{e.g.} a forklift can be partially blocking the reference trajectory). During execution, our vehicle selects a new trajectory $\Traj$ from a predefined motion primitive library $\library = \{\Traj_k\}_{k=0}^K$ to follow such that it remains probabilistically safe under current disturbance while minimizing the similarity metric $J_{sim}$. Eq.~\ref{eq:overall_objective} defines our replanning objective, with the probabilistic safety constraint represented by trajectory-relative safety tube (Eq.~\ref{eq:tube_def}).
\begin{equation}
    \Traj^{*} = \argmin_{\Traj \in \library}~J_{sim}(\Traj, \TrajRef),~~s.t.~~\tube_\Traj \in \FreeSpace \label{eq:overall_objective} \\
\end{equation}

\section{Approach} 
\label{sec:approach}

This section details the overall approach to solve our replanning objective in Eq. \ref{eq:overall_objective}. 

We propose a \textit{offline-online} framework that uses a library of offline precomputed tubes and an online disturbance estimator to select motion primitives based on uncertainty in real-time. Fig. \ref{fig:approach} describes the overall process. The framework is generalizable to other mobile robot platforms, where we focus on Unmanned Aerial Vehicles (UAVs) in this paper.

\textit{Offline}, given a motion primitive library, we generate a lookup table (LUT) of safety margins for each motion primitive at discretised levels of disturbance. For every motion primitive and disturbance level, we estimate a safety tube as in Eq.~\ref{eq:tube_def}, where we calculate the tube margin by first generating possible trajectories using Monte Carlo simulations (Sec.~\ref{sec:MC}), then fitting a tube (Sec.~\ref{sec:Tubes}).  
\textit{Online}, we estimate the disturbance variance (Sec.~\ref{sec:dist_model}). Then, given a reference trajectory, the variance is used to select the optimal collision-free motion primitive with collision check based on the precomputed tubes from LUT (Sec.~\ref{sec:Margin Lookup}). Disturbance mean is set to zero and we use the variance to capture varying disturbances.

\subsection{Offline: Generating Margin Lookup Table (LUT)}
\subsubsection{Simulating Possible Trajectories} 
\label{sec:MC} 
Given a set of disturbance standard deviations $\bm{\sigma} = \{\sigma_g\}_{g=0}^G$ and a motion primitive library $\library$ , we run $N_{mc}\times K \times G$ Monte Carlo simulations to collect a set of possible vehicle trajectories $\Data{\mathcal{L},\bm{\sigma}}$. We use Monte Carlo simulations as it allows us to generate predictions for arbitrary dynamics model, underlying disturbance model and controller, without any assumptions of their structure.
The Monte-Carlo simulated vehicle follows a motion primitive $\Traj_k$ with controller $\pi$ whilst under nominal model dynamics $\hat{\fv}$ with a disturbance sampled from $\mathcal{N}(0,\sigma_g)$ every disturbance period and an initial state distribution defined by an operator $N_{mc}$ times to generate a set $\Data{\Traj_k,\sigma_g}$. 
\\
\subsubsection{Tube Fitting}
\label{sec:Tubes}

We find the optimal margin radius $\theta$ that define a minimum-volume tube that encloses predicted states in $\Data{\Traj_k,\sigma_g}$.
We discretize trajectory to segments by time and fit a zero-mean normal distribution to cross-track error of collected points in each segment. For each segment, we find the fitted interval (e.g., $2\sigma$) that corresponds to desired confidence bound $1-\epsilon$. The maximum fitted interval over all segments is then added to LUT, which can be queried by $\theta_{\Traj_k, \sigma_g} = LUT\big[\Traj_k, \sigma_g\big]$. 
\subsection{Online: Replanning with Adaptive Margins}

\subsubsection{Temporal Disturbance Modeling}
\label{sec:dist_model}
We estimate the current disturbance term $\dv_t$ by learning a disturbance model $\bar{\gv}(\bm{\Traj}) \rightarrow \sigma_{\bar{\gv}}$.
Disturbance is the difference between predicted and actual dynamics $\bar{\gv} = \dot{\xv}_t - \hat{\fv}(\xv_{t-1}, \uv_{t-1})$.
In this paper, we set $\sigma_{\bar{g}}^2$  as a moving variance on observed disturbance, and assume all primitives share the same time-varying disturbance. However, any disturbance modeling method that provides a disturbance variance can be used.\\
\subsubsection{Margin Lookup and Trajectory Selection}
\label{sec:Margin Lookup}

\begin{algorithm}[ht]
 \small
\SetAlgoLined
\KwInput{Reference Trajectory $\Traj_{ref}$, Disturbance variance $\sigma_g$, Free Space $\FreeSpace$, Margin Lookup Table $LUT$, Motion Primitive Library $\library$}

\textbf{Initialize: }$min\_cost = \infty,~\Traj^*$\;
\ForEach{primitive $\Traj$ in library $\library$}{
    $\tube_{\Traj} = LUT\big[\Traj, \sigma_g\big]$ \Comment{\footnotesize lookup tube}
    \If{$\tube_{\Traj} \in \FreeSpace$}{
        $cost = J_{sim}(\Traj, \Traj_{ref})$\;
        \If{$cost < min\_cost$}{
            $\Traj^* = \Traj$, $min\_cost = cost$\;
        }
    }
}
\Return{$\Traj^*$}
\caption{\small Trajectory Selection with Adaptive Margins}
\label{alg:online_planning}
\end{algorithm}

Given estimated disturbance variance $\sigma_{\bar{g}}^2$, we first query the associated safety margins for each primitive $\Traj_k$ in the library by using the disturbance $\sigma_g = \lceil \sigma_{\bar{g}} \rceil$.
 The collision-free (given respective safety margin) primitive with minimum trajectory cost is then chosen, here we use L1 norm. 
   \begin{equation}
     J_{sim}(\Traj, \TrajRef) = \frac{1}{t_f}\int^{t_f}_0\|\Traj(t) - \Traj_{ref}(t)\|
     \vspace{-5pt}
 \end{equation}
 The margin lookup and trajectory selection step is shown in Alg. \ref{alg:online_planning}.

\subsection{Implementation}
In this paper, we replan at 5Hz with a motion primitive library of constant angular and linear velocity. Figure \ref{fig:lut_rw} shows a sample lookup table used in maze trials. After parallelization, this 22-trajectory, 9-variance LUT with 1000 Monte Carlo simulations took $\sim$230s to produce and uses 1.1kB of space. 
 A position and velocity PID controller computes desired attitude. We estimate body-x and y disturbance ($g_x$, $g_y$) in acceleration. The larger between $\sigma_{\bar{g_x}}$ and $\sigma_{\bar{g_y}}$ is used for lookup. Our nominal model is modified from \cite{kamelmpc2016} to assume linear mapping between desired and normalized thrust, and desired yaw-rate is actual yaw-rate.

\section{Simulated Experiments}
\begin{figure}[t]
  \centering
  \includegraphics[width=1\textwidth]{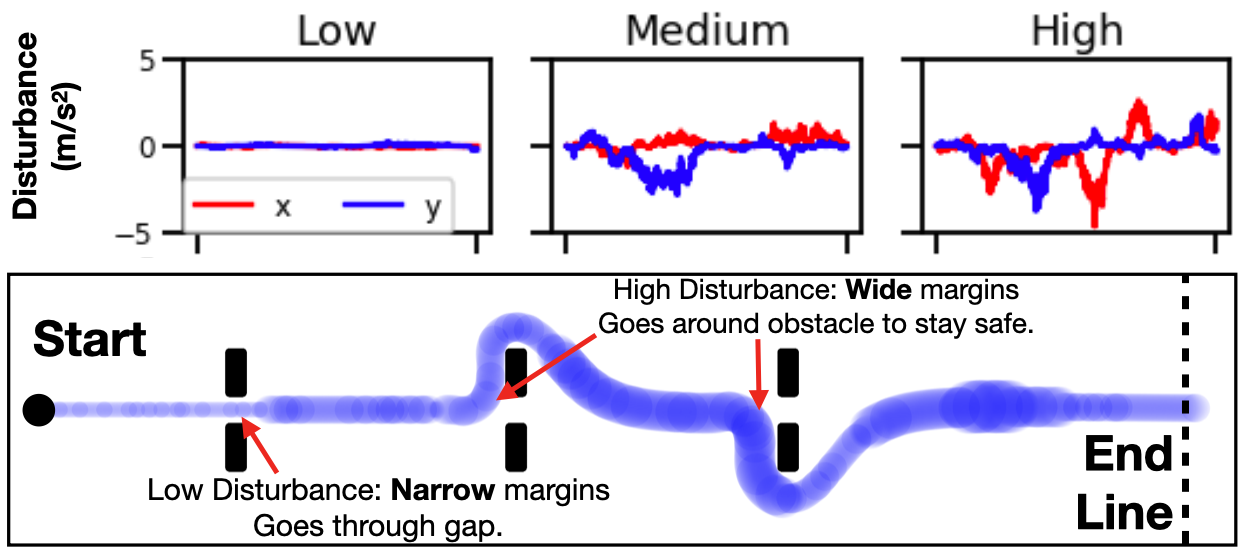}
  \caption{Simulated setup. (\textit{Top}) Samples of Dryden wind gusts. (\textit{Bottom}) Obstacle configuration. Shaded blue region shows adaptive margins over one sample trial, where vehicle goes through first gap when margin is small and goes around obstacle when disturbance is high.}
  \label{fig:sim_setup}
\end{figure}
We compare the performance of our adaptive margins to baseline static margins under increasing magnitude of disturbances. We validate the transferiblity of our method to realistic wind situations by using wind from the Dryden gust model \cite{beardMclain2012}, often used by the Department of Defense.

Fig.~\ref{fig:sim_setup} shows map and samples of wind sequences. The vehicle is given a straight line path and a trial is successful if vehicle reaches goal line without crashing. We use the same tube library at fixed disturbance variance (\textit{Static: Low} at $0m/s^2$, \textit{Static: High} at $5m/s^2$) as our two baselines. For our margin lookup, we use the zero-centered standard deviation of the future 9s sequence of disturbance. Each data point is an averaged result over 10 trials in randomized wind sequences.

Table~\ref{tab:gt_dryden} shows the performance of different margin methods, under increasing levels of turbulence.
At all turbulence level, our adaptive margins dynamically computes best margin given current information, resulting in overall lower cost than  \textit{Static: High}, while maintaining 100\% safety. The rightmost column of Table~\ref{tab:gt_dryden} shows the percentage of time robot is within margin. The results confirm our precomputed margin transfers to realistic wind conditions, with worst within-margin percentage being $99.68\%$ at high turbulence, likely due to suddenly higher gusts pushing the vehicle out very briefly. In contrast, \textit{Static: Low} only stays within margin $\sim80\%$ of time, leaving it more vulnerable to collisions with $3/10$ trials failing.

\begin{table}[ht]
\vspace{-5pt}
  \footnotesize
  \begin{tabular}{c|c|c|c|c}
  \hline
  \textbf{\begin{tabular}[c]{@{}c@{}}Turb. \\ Level\end{tabular}} &
  \textbf{\begin{tabular}[c]{@{}c@{}}Margin \\ Type\end{tabular}} &
    \textbf{\begin{tabular}[c]{@{}c@{}}Success \\ (\%)\end{tabular}} &
    \textbf{\begin{tabular}[c]{@{}c@{}}Avg. Distance\\ to Reference (m)\end{tabular}} &
    \textbf{\begin{tabular}[c]{@{}c@{}}Within \\ Margin (\%)\end{tabular}} \\ \hline
  Low &
    \begin{tabular}[c]{@{}c@{}}Adaptive\\ Static: Low\\ Static: High\end{tabular} &
    \begin{tabular}[c]{@{}c@{}}\textbf{100}\\ \textbf{100}\\ \textbf{100}\end{tabular} &
    \begin{tabular}[c]{@{}c@{}}\textbf{0}\\ \textbf{0}\\ 3.84$\pm$0.03\end{tabular} &
    \begin{tabular}[c]{@{}c@{}}\textbf{100}\\ \textbf{100}\\ \textbf{100}\end{tabular} \\ \hline
  Med &
    \begin{tabular}[c]{@{}c@{}}Adaptive\\ Static: Low\\ Static: High\end{tabular} &
    \begin{tabular}[c]{@{}c@{}}\textbf{100}\\ 90\\ \textbf{100}\end{tabular} &
    \begin{tabular}[c]{@{}c@{}}\textbf{1.61$\pm$0.46}\\ 0\\ 3.86$\pm$0.02\end{tabular} &
    \begin{tabular}[c]{@{}c@{}}99.96$\pm$0.12\\ 87.71$\pm$12.16\\ \textbf{100}\end{tabular} \\ \hline
  High &
    \begin{tabular}[c]{@{}c@{}}Adaptive\\ Static: Low\\ Static: High\end{tabular} &
    \begin{tabular}[c]{@{}c@{}}\textbf{100}\\ 70\\ \textbf{100}\end{tabular} &
    \begin{tabular}[c]{@{}c@{}}\textbf{2.37$\pm$0.32}\\ 0\\3.86$\pm$0.04\end{tabular} &
    \begin{tabular}[c]{@{}c@{}}99.68$\pm$0.59\\ 80.18$\pm$13.77\\ \textbf{100}\end{tabular} \\ \hline
  \end{tabular}
  \caption{\small Performance of different margin methods under simulated gusts. Our adaptive margins results in lower cost while maintaing 100\% safety, better balancing safety and performance. 
  \vspace{-4.0mm}}
  \label{tab:gt_dryden}
  \end{table}

\section{Real-World Experiments} 
\label{sec:experiments}

We test with real flights three hypotheses: Our adaptive margin method can (\textbf{H1}) adjust margins in real-time, (\textbf{H2}) better balance safety and performance than baseline static margins in tight spaces, (\textbf{H3}) and under time-varying disturbances.

\textbf{Experimental Platform:} We use a custom-built quadrotor with an Intel T265 Tracking Camera for state estimation. All computation is done onboard using a Jetson TX2. An offline map is given for all trials.

\begin{figure}[tbp]
    \centering
    \includegraphics[width=1\textwidth]{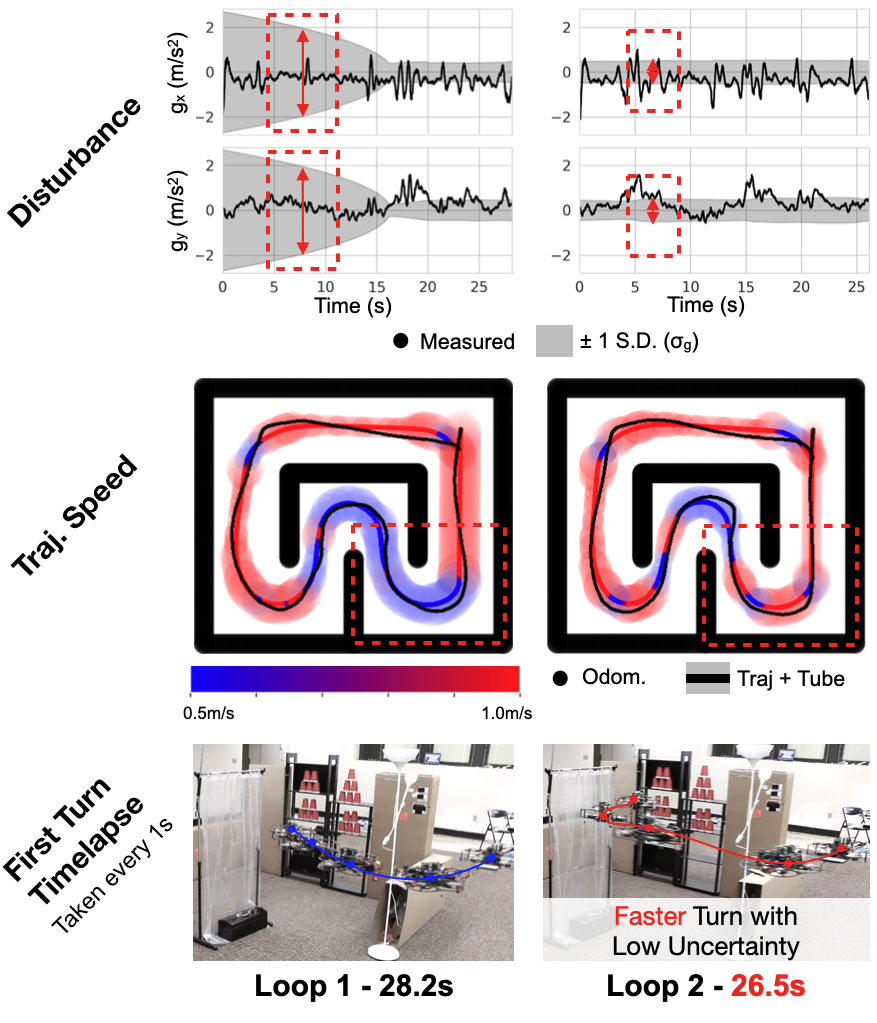}
    \caption{\small \textbf{E1 - Real-time adaptation to disturbance uncertainty}: In this experiment, our quadrotor completes two loops around the maze with disturbance uncertainty decreasing over time. Safety margins adaptively decrease with lower uncertainty. At bottom-right corner, system picked the faster trajectory in the second loop to complete the turn. Actual odometry (black) stays within tubes in all trials. Across two loops, average trajectory cost decreased from 0.249 to 0.218, and completion time from 28.2s to 26.5s. 
    \vspace{-2.0mm}}
    \label{fig:adaptive-trials}
  \end{figure}
\begin{wrapfigure}{h}{0.15\textwidth}
    \centering
    \vspace{-10pt}
    \includegraphics[width=\textwidth]{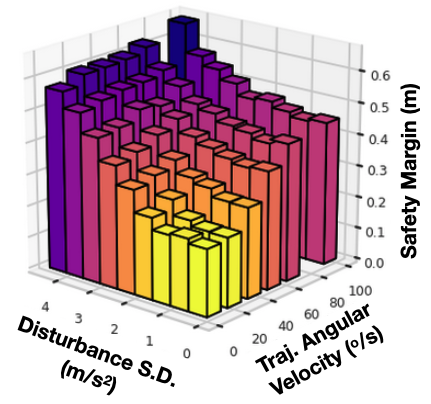}
    \caption{\small 0.5m/s lookup table used in \textbf{E1}/\textbf{E2}.} 
    \vspace{-10pt}
    \label{fig:lut_rw}
\end{wrapfigure}

 \resizebox{0.49\textwidth}{!}{\textbf{E1 - Real-time adaptation to disturbance uncertainty (H1):}}
To validate the real-time nature of our method, we set up an indoor maze environment with tight corridors (1.5m) and $180^\circ$ hairpin turns, as shown in Fig.~\ref{fig:adaptive-trials}. The configuration is difficult to navigate with potential crashes around turns due to undercutting or overshooting. The narrow environment also causes very high safety margins to render no solution. A collision-free 1m/s global plan is given. We show the impact of decreasing margin size,by precomputing margins for two sets of motion primitive library, one at $0.5m/s$, another at $1m/s$. 
Fig.~\ref{fig:lut_rw} shows this experiment's LUT for the 0.5m/s motion primitive library. As expected, computed margins increase with higher uncertainty and higher angular velocity. 

To show effects of varying disturbance uncertainty, the first loop around the maze is initialized with a high disturbance standard deviation at $\sigma_{\bar{g}}=3.0\text{m/s}^2$, and the second loop with the first loop's final standard deviation. 
Fig. \ref{fig:adaptive-trials} compares the two loops, with narrower margins, lower completion time and cost as disturbance uncertainty decreases. For example, in the lower right turn, system picked the faster trajectory margins in the second loop as its margins decreased with lower uncertainty,  Table~\ref{table:runtime} validates real-time performance, with trajectory selection taking 26ms and margin lookup only taking 0.02 ms for a library of 22 motion primitives.\\

\begin{table}[hbtp]
\footnotesize
\caption{\small \textbf{E2 - Comparison to static margin methods in narrow environment.}
Over ten trials, our method computes more optimal plans while remaining safe. $^*$ indicates over successful trials, and $^\#$ is the nominal completion time of 18.4s.
\vspace{-4.0mm}}
\label{table:compare_baseline}%
{\begin{tabular}{l|l|l|l}
\hline
\textbf{\begin{tabular}[c]{@{}l@{}}Margin Type\\ {[0.5m/s, 1m/s]}\end{tabular}}                                                                               
& \textbf{\begin{tabular}[c]{@{}l@{}}Success $\%$\\ 10 Trials\end{tabular}} 
& \textbf{\begin{tabular}[c]{@{}l@{}}Completion \\Time (s)$^\#$ \end{tabular}} 
    & \textbf{\begin{tabular}[c]{@{}l@{}}Avg. Distance\\to Reference (m) $^\#$ \end{tabular}} \\ \hline
\textbf{\begin{tabular}[c]{@{}l@{}}Adaptive\end{tabular}    }                                                                   
    &       \textbf{100\%}                
    &       \begin{tabular}[c]{@{}l@{}}\textbf{\textit{26.5 $\pm$}}  \\\textbf{\textit{0.6}}\end{tabular}                                                              
    &  \begin{tabular}[c]{@{}l@{}}\textbf{\textit{0.222$\pm$}}  \\ \textbf{\textit{0.006}}\end{tabular}                            \\ \hline
\begin{tabular}[c]{@{}l@{}}Conservative\\ {[}40cm, 2m{]}\end{tabular}           
    &        \textbf{100\%}               
    &                   \begin{tabular}[c]{@{}l@{}}39.9  $\pm$\\0.3\end{tabular}                                                           
    &  \begin{tabular}[c]{@{}l@{}}0.391  $\pm$\\ 0.002\end{tabular}                         \\ \hline

\begin{tabular}[c]{@{}l@{}}Handtuned\\ {[}15cm, 15cm{]}\end{tabular} 
    &          30\%              
    &                    \begin{tabular}[c]{@{}l@{}}\textbf{24.5 $\pm$}  \\\textbf{0.3$^*$}\end{tabular}                                                   
    &     \begin{tabular}[c]{@{}l@{}}\textbf{0.193 $\pm$}\\ \textbf{0.005$^*$}\end{tabular}                           \\ \hline
    \begin{tabular}[c]{@{}l@{}}Handtuned\\ {[}20cm, 20cm{]}\end{tabular} 
        &          90\%              
        &                     \begin{tabular}[c]{@{}l@{}}24.9 $\pm$  \\0.4\end{tabular}                                                           
        &  \begin{tabular}[c]{@{}l@{}}0.195 $\pm$\\ 0.002\end{tabular}                          \\ \hline
\end{tabular}}
\end{table}

\begin{table}[tbp]
    \footnotesize
    \begin{tabular}{l|lll}
        \hline
            \textbf{Procedure}                                                                         & \textbf{Runtime(ms)} & \textbf{CPU Thread \%} \\ \hline
  \begin{tabular}[c]{@{}l@{}}Traj. Selection\\  - Margin Lookup\end{tabular} & \begin{tabular}[c]{@{}l@{}} 26\\  - 0.02\end{tabular}       & \begin{tabular}[c]{@{}l@{}}31.6\\ - \end{tabular}        \\ \hline
    \end{tabular}
    \caption{\small \textbf{E1} Runtime Statistics on Jetson TX2 
    \vspace{-4.0mm}}
    \label{table:runtime}
\end{table}

  \begin{figure*}[]
    \centering
    \includegraphics[width=0.9\textwidth]{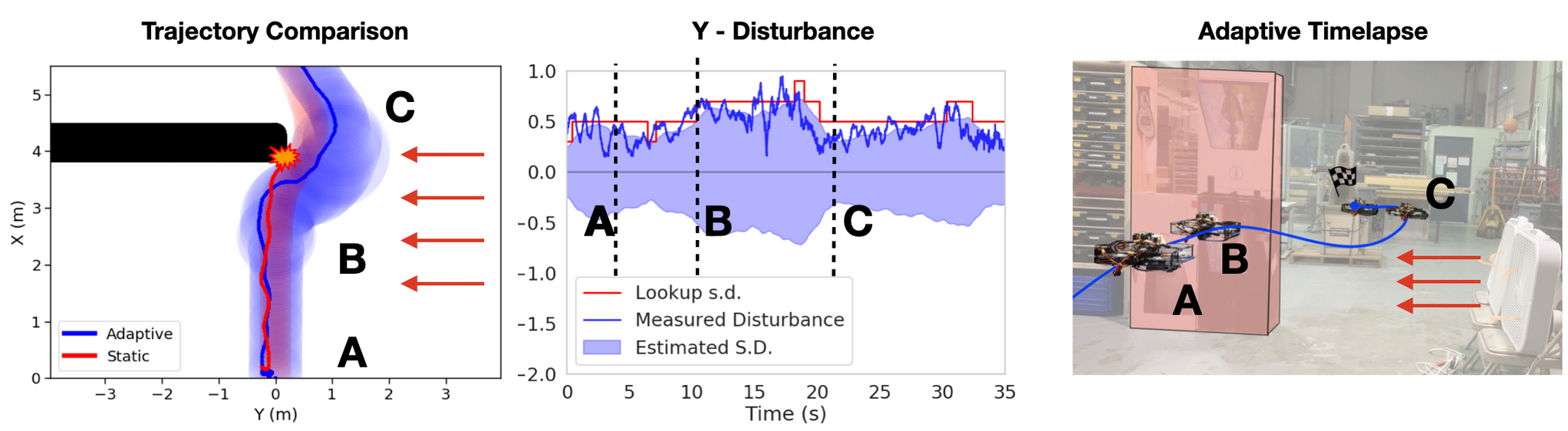}
    \caption{\small \textbf{E3 - Adapting to Time-varying Disturbances}. Our adaptive margin (blue) increases when under disturbance (\textbf{B}), enabling the robot to successfully avoid the obstacle. In contrast, quadrotor with static margin (red) crashes under wind. Red line in center figure indicates disturbance standard deviation used for margin lookup. Solid line shows actual odometry, where all robot states is within our adaptive margins.
    \vspace{-4.0mm}}
    \label{fig:rw_4}
\end{figure*}

\textbf{E2 - Baseline comparisons in tight spaces (H2):} We compare our adaptive approach to static margin methods with the same setup as E1. A trial is successful if the vehicle successfully completes one maze loop without crashing. We use three sets of static margins as baselines: one conservative using high bounded disturbance from \cite{fridovich-keil_planning_2018}, and two handcrafted static margins tuned with other global plans. Our adaptive margins are hot-started at $\sigma_{\bar{g}}=0.45\text{m/s}^2$ from E1. Table \ref{table:compare_baseline} compares success rate, completion time and cost, with our adaptive margins enabling higher-performance flight than conservative margins, while staying 100\% safe whereas handcrafted margins did not. 
The results display the difficulty of handcrafting safety margins, where a difference of 5cm leads to 60\% more crash.

\textbf{E3 - Adapting to Time-varying Disturbance (H1+H3)}:
We validate the performance of our proposed framework under time-varying disturbances by testing its ability to avoid obstacles in windy environment. The quadrotor is given a straight line path with an obstacle straight ahead and wind blowing laterally (in y-direction) towards the obstacle. Fans are not included in map. Fig.~\ref{fig:rw_4} shows experimental setup and our vehicle's planned trajectories with computed margins when flying through wind. We use a static margin handtuned with no wind as baseline. Red line on y-disturbance plot shows disturbance standard deviation used to lookup margins. 
The vehicle starts at a low disturbance at (A). 
At (B), when vehicle encounters wind and experience a high y-disturbance, the margins dynamically increase in size to move the quadrotor away from danger. Note, right after (B), the robot moves 30cm to the left which would have been at the edge of previous tube if it had not adapted. 
For most of the time quadrotor is in the wind field between (B) and (C), lookup s.d. is $0.5m/s^2$, with a jump to $0.7m/s^2$ as the quadrotor moves towards the last fan while avoiding the obstacle.  At (C), vehicle successfully avoids the obstacle and decreases its margin as disturbance has lowered.  In contrast, the quadrotor with the fixed margin leaves its tube and crashes. \\
Our adaptive margins allow us to stay high-performance (smaller margins) when disturbance is low, and safe when disturbance is high. The instantaneous increase of margins to wind disturbance validates the real-time performance.

\section{Conclusion and Discussion}
\label{sec:conclusion}
In this paper, we present a real-time framework for \textit{adapting} safety margins \textit{on-the-fly} to compute safe plans without overly sacrificing performance. Central to this approach, is the precomputation of safety margins at multiple disturbance levels for a motion primitive library. We validated with extensive simulated and real flights that our adaptive margins significantly increase performance while maintaining safety in tight spaces and under time-varying disturbances than static margin methods.

We find multiple directions for future work. We are interested in coupling our adaptive framework with SOS optimization as used in \cite{majumdar_funnel_2017,singh_robust_2019, gahlawat2020l1gp}, to precompute margins for \textit{multiple} levels of worst-case disturbances. This removes our need to define the Monte Carlo disturbance model and enables better handling of disturbance bias. We can also extend our 1-step reactive planning framework to multiple steps for longer horizon planning \cite{yang2020}.
Finally, we are working on including rigorous safety guarantee when transitioning between primitives, taking inspirations from \cite{fridovich-keil_planning_2018, majumdar_funnel_2017, lakshmanan2020safe}.

\section*{ACKNOWLEDGMENT}
This work was supported by the Croucher Foundation, the Office of Naval Research (Grant $\#$N00014-18-R-SN05), Department of Energy Grant DE-EE0008463, and the Waibel scholarship.
The authors would like to thank Christoph Endner and Tino Fuhrmann for construction of the platform, and Brady Moon for Dryden model implementation.

\footnotesize{
\bibliographystyle{IEEEtran}
\bibliography{reference}
}

\end{document}